\documentclass[conference]{IEEEtran}
\IEEEoverridecommandlockouts
\usepackage{cite}
\usepackage{amsmath,amssymb,amsfonts}
\usepackage[english]{babel}
\usepackage{algorithmic}
\usepackage{graphicx}
\usepackage{textcomp}
\usepackage{balance}
\usepackage{booktabs}
\usepackage{multirow}
\usepackage{enumitem}
\usepackage[colorlinks=true, linkcolor=blue, urlcolor=blue]{hyperref}
\usepackage{caption}
\addto\extrasenglish{%
  
}
\usepackage[T1]{fontenc}
\usepackage{comment}
\usepackage{float} % to use [H] option
\usepackage{xcolor}
\usepackage{acronym}
\usepackage{bm}

\def\BibTeX{{\rm B\kern-.05em{\sc i\kern-.025em b}\kern-.08em
    T\kern-.1667em\lower.7ex\hbox{E}\kern-.125emX}}  

\begin{document}

\title{A Systematic Benchmark of GAN Architectures for MRI-to-CT Synthesis}

\author{
\IEEEauthorblockN{
Alessandro Pesci\IEEEauthorrefmark{1}\IEEEauthorrefmark{5},
Valerio Guarrasi\IEEEauthorrefmark{1}\IEEEauthorrefmark{5},
Marco Alì\IEEEauthorrefmark{3},
Isabella Castiglioni\IEEEauthorrefmark{3}\IEEEauthorrefmark{4},
Paolo Soda\IEEEauthorrefmark{1}\IEEEauthorrefmark{2}
}

\IEEEauthorblockA{
\IEEEauthorrefmark{1}Unit of Artificial Intelligence and Computer Systems, Department of Engineering,\\
Università Campus Bio-Medico di Roma, Italy
}

\IEEEauthorblockA{
\IEEEauthorrefmark{2}Department of Diagnostics and Intervention, Biomedical Engineering and Radiation Physics,\\
Umeå University, Umeå, Sweden
}

\IEEEauthorblockA{
\IEEEauthorrefmark{3}CDI Centro Diagnostico Italiano S.p.A., Milan, Italy
}

\IEEEauthorblockA{
\IEEEauthorrefmark{4}University of Milan-Bicocca, Milan, Italy
}

\IEEEauthorblockA{
\IEEEauthorrefmark{5}Alessandro Pesci and Valerio Guarrasi contributed equally to this work.
}

\IEEEauthorblockA{
{alessandro.pesci, valerio.guarrasi, p.soda}@unicampus.it
}
}
\maketitle

\begin{abstract}
The translation from Magnetic resonance imaging (MRI) to Computed tomography (CT) has been proposed as an effective solution to facilitate MRI-only clinical workflows while limiting exposure to ionizing radiation.
Although numerous Generative Adversarial Network (GAN) architectures have been proposed for MRI-to-CT translation, systematic and fair comparisons across heterogeneous models remain limited.
We present a comprehensive benchmark of ten GAN architectures evaluated on the SynthRAD2025 dataset across three anatomical districts (abdomen, thorax, head-and-neck). All models were trained under a unified validation protocol with identical preprocessing and optimization settings. Performance was assessed using complementary metrics capturing voxel-wise accuracy, structural fidelity, perceptual quality, and distribution-level realism, alongside an analysis of computational complexity.
Supervised Paired models consistently outperformed Unpaired approaches, confirming the importance of voxel-wise supervision. Pix2Pix achieved the most balanced performance across districts while maintaining a favorable quality-to-complexity trade-off. Multi-district training improved structural robustness, whereas intra-district training maximized voxel-wise fidelity. 
This benchmark provides quantitative and computational guidance for model selection in MRI-only radiotherapy workflows and establishes a reproducible framework for future comparative studies. To ensure the reproducibility of
our experiments we make our code public, together with the overall results, at the following link: \url{https://github.com/arco-group/MRI_TO_CT.git}

\end{abstract}
\begin{IEEEkeywords}
Generative AI, GANs, MRI, CT, image-to-image translation, Benchmarking 
\end{IEEEkeywords}

\section{Introduction}
Computed Tomography (CT) and Magnetic Resonance Imaging (MRI) represent fundamental imaging modalities in both clinical diagnostics and radiotherapy planning, as they provide complementary information on the anatomical and functional status of the patient. While MRI offers superior soft-tissue contrast without exposure to ionizing radiation, making it particularly suitable for target delineation and longitudinal follow-up~\cite{bib:mri}, CT provides quantitative information on tissue electron density that is indispensable for accurate radiation dose calculation~\cite{bib:ct}.
However, the use of CT entails exposure to X-ray radiation, which is associated with dose-dependent biological risks, especially in vulnerable populations or in repeated imaging scenarios, as well as increased costs and complexity of clinical workflows~\cite{bib:ct_risk}. In this context, the generation of synthetic CT images from MRI has emerged as a promising solution to enable MRI-only radiotherapy paradigms, in which treatment planning is performed without acquiring a planning CT scan.
This approach has the potential to reduce radiation exposure, eliminate additional imaging sessions, and simplify image co-registration procedures~\cite{tronchin2025latentaugment}. Numerous studies have addressed the MRI-to-CT translation task using image-to-image translation techniques based on Generative Adversarial Networks (GANs), which currently represent the predominant approach in the literature due to their ability to learn complex cross-modal mappings and generate realistic synthetic images~\cite{bib:MRITOCTComparative}. Among the various GAN-based frameworks proposed for this task, Pix2Pix and CycleGAN have emerged as the most widely adopted architectures, commonly serving as supervised and unsupervised reference models, respectively, and frequently employed as benchmark baselines for comparative evaluation in MRI-to-CT synthesis studies\cite{bib:mri_to_ct_review_2}.
Despite the widespread adoption of GAN-based approaches for MRI-to-CT synthesis, the literature still lacks a systematic and standardized comparison across heterogeneous GAN architectures, encompassing both quantitative performance assessment and computational complexity analysis. In this work, we address this gap by proposing:
\begin{itemize}
    \item a standardized and reproducible benchmark framework for MRI-to-CT translation that includes ten GAN architectures representative of different methodological families;
    \item a quantitative analysis of model performance using image fidelity, structural similarity, and perceptual quality metrics, evaluated under identical data splits, preprocessing pipelines, and training protocols to ensure fairness;
    \item an evaluation of model computational complexity in terms of number of parameters and computational cost, providing insight into the trade-off between performance and efficiency.
\end{itemize}

\section{Materials and Methods}

\begin{figure}[t]
    \centering
    \includegraphics[width=\linewidth]{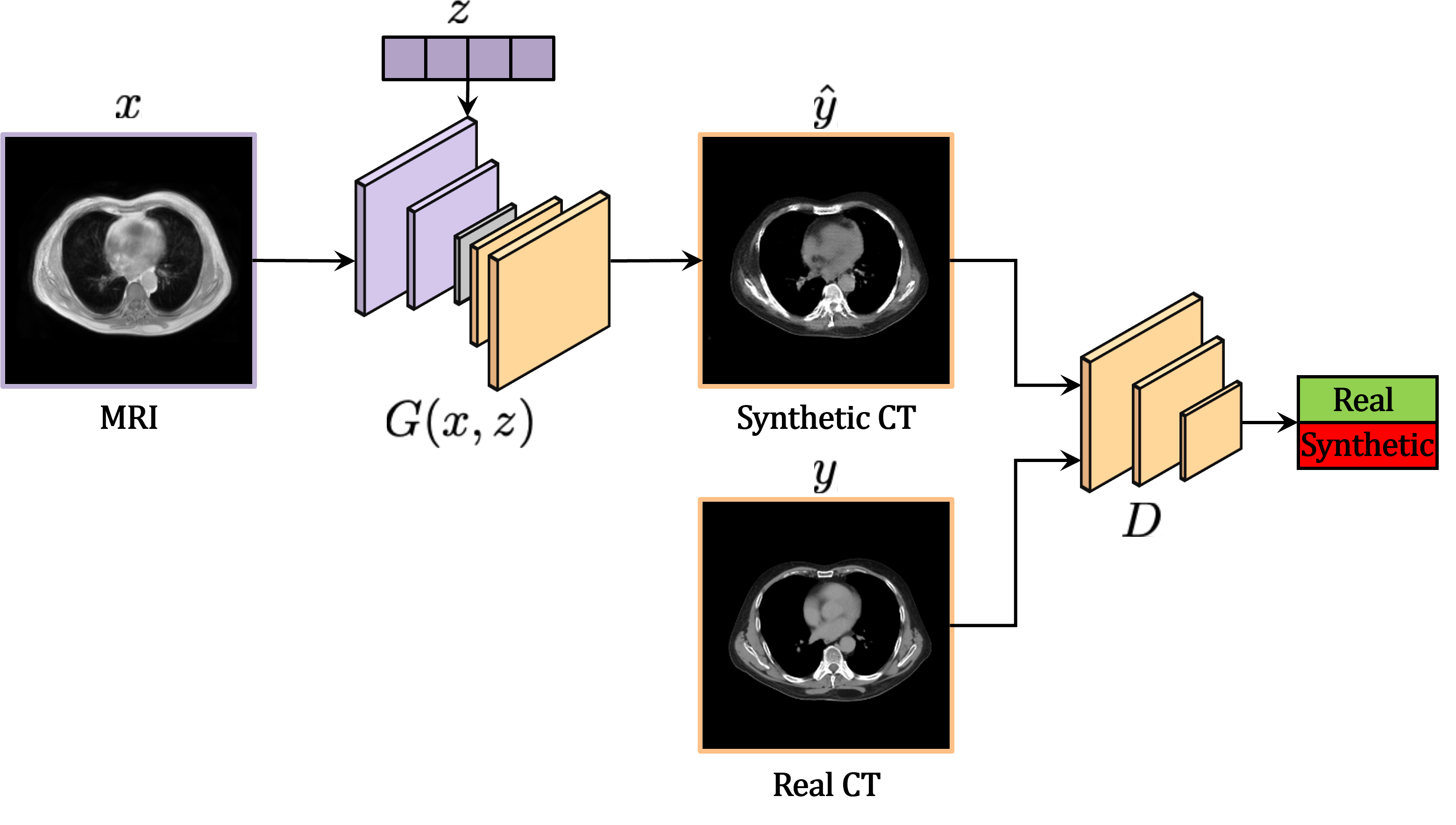}
    \caption{
    General GAN-based framework for MRI-to-CT image translation.
    }
    \label{fig:mri2ct_gan_framework}
\end{figure}

\subsection{Dataset and Pre-processing}

In this study, we used the SynthRAD2025 dataset, released as part of the SynthRAD2025 Grand Challenge~\cite{thummerer2025synthrad2025}. The dataset includes  513 patients with paired MRI and CT 3D scans. The paired examinations are organized into three anatomical regions: abdomen (AB, 175 patients), thorax (TH, 182 patients), and head-and-neck (HN, 156 patients). All volumes are provided with voxel-wise spatial correspondence between MRI and CT, enabling supervised training.\\
Pre-processing started from the 3D CT and MRI volumes of the SynthRAD2025 dataset.
To reduce acquisition artifacts and terminal redundancies, the last 10\% of slices along the \(z\)-axis were removed. Subsequently, constant or fully NaN slices were jointly removed from CT and MRI volumes (same slice indices) to preserve alignment. These slices were identified using a zero peak-to-peak intensity criterion (max–min \(=0\)), computed slice-wise while ignoring NaN values.
After volumetric filtering, each volume was decomposed into paired 2D slices with identical slice indices.A further quality-based filtering step was applied to discard slices dominated by background. Specifically, if the proportion of pixels equal to \(-1000\)  HU exceeded a district-specific threshold (AB: 43\%, TH: 33\%, HN: 30\%), the slice was excluded from the dataset. These thresholds were empirically determined to remove non-informative slices while preserving relevant anatomical content. This additional filtering prevents the inclusion of slices lacking meaningful structures, which could otherwise introduce bias during training, destabilize the optimization process, and negatively affect the learning of anatomically consistent MRI-to-CT mappings. During data loading, additional pre-processing steps were applied on-the-fly:
(i) zero-padding to obtain square images,
(ii) CT windowing with district-specific window center/width (AB: 50/400, TH: 40/400, HN: 40/80),
(iii) robust normalization to \([0,1]\),
(iv) isotropic resizing to \(256 \times 256\)  using bilinear interpolation,
(v) data augmentation applied only during training and synchronously to MRI and CT slices.

\subsection{GAN Formulation}

We address the image-to-image translation problem of synthesizing CT images from MRI data (Figure~\ref{fig:mri2ct_gan_framework}). Let \(X\) denote the MRI domain and \(Y\) the CT domain. The objective is to learn a mapping that, given an MRI image \(x \in X\), generates a synthetic CT image \(\hat{y} \in Y\) that is both visually realistic and quantitatively consistent with the corresponding real CT image \(y \in Y\).

The transformation is modeled by a generator:
\[
G : X \times Z \rightarrow Y,
\]
where \(Z\) denotes a latent space capturing stochastic variability. In deterministic supervised settings, the mapping reduces to \(G: X \rightarrow Y\).

The synthetic CT image is generated as:
\[
\hat{y} = G(x, z).
\]

A discriminator \(D\) is trained to distinguish real CT images from generated ones. The optimization follows the standard min--max adversarial objective, where the generator aims to produce indistinguishable synthetic CT images while the discriminator attempts to classify them correctly. Different GAN variants are obtained by modifying architectural components and loss formulations (e.g., reconstruction, cycle-consistency, KL-divergence, or domain-classification losses). 

\subsection{GAN Architectures}
\label{Gan Architectures}
The analyzed models were grouped into five families according to their supervision strategy and structural design~\cite{bib:ClassificationGAN}:

\begin{itemize}

\item \textbf{Paired (Supervised) Models:}
This family includes conditional GANs trained on aligned image pairs \((x,y)\), leveraging voxel-wise supervision. Pix2Pix~\cite{bib:Pix2Pix} learns a deterministic mapping \(G: X \rightarrow Y\) using an adversarial loss combined with an $\ell_1$ reconstruction loss to enforce intensity fidelity and stabilize optimization. BicycleGAN~\cite{bib:BicycleGAN} extends this framework to multimodal generation by introducing a latent variable and an encoder network. In addition to adversarial and reconstruction losses, it incorporates Kullback--Leibler divergence regularization and a latent consistency loss, enabling one-to-many mappings while preserving structural coherence.

\item \textbf{Cycle-Based (Unpaired) Models:}
Cycle-Based models (CycleGAN~\cite{bib:CycleGAN}, DualGAN~\cite{bib:DualGAN}, DiscoGAN~\cite{bib:DiscoGAN}) perform unpaired image translation by learning two generators, \(G: X \rightarrow Y\) and \(F: Y \rightarrow X\), along with domain-specific discriminators. In addition to adversarial losses, these models introduce a cycle-consistency constraint to reconstruct the original image after bidirectional translation, thereby preserving semantic and structural information without paired supervision. CycleGAN includes an identity loss to improve color and intensity preservation; DualGAN adopts a Wasserstein formulation to enhance training stability; and DiscoGAN emphasizes explicit bidirectional reconstruction to mitigate mode collapse.

\item \textbf{Shared Latent Space Models:}
This family (CoGAN~\cite{bib:CoGAN}, UNIT~\cite{bib:UNIT}, MUNIT~\cite{bib:MUNIT}) assumes that both domains can be embedded into a shared latent representation. CoGAN trains two GANs with partially shared generator weights to promote implicit semantic alignment. UNIT combines variational autoencoder (VAE) and GAN objectives under a shared Gaussian latent space, integrating reconstruction and KL-divergence losses with adversarial learning. MUNIT further decomposes the latent space into shared content and domain-specific style components, enabling multimodal translation through content--style recombination.

\item \textbf{Multi-Domain Model:}
StarGAN~\cite{bib:StarGAN} performs translation across multiple domains using a single conditional generator \(G(x,c)\), where \(c\) encodes the target domain. The discriminator incorporates both adversarial and domain classification branches. Training combines adversarial loss (with gradient penalty), domain classification loss, and reconstruction loss to enable scalable and controllable multi-domain translation within a unified framework.

\item \textbf{Domain Adaptation Model:}
PixelDA~\cite{bib:PixelDA} addresses unsupervised domain shift by learning a pixel-level transformation from a source to a target domain. The architecture comprises a generator, discriminator, and auxiliary task classifier. In addition to adversarial alignment, task consistency and content similarity losses are employed to preserve semantic structure. Unlike cycle-based approaches, PixelDA performs unidirectional adaptation without explicit cycle-consistency constraints.

\end{itemize}

All selected architectures were implemented following their original formulations, preserving the core loss functions and training objectives to ensure a fair and representative comparison across paradigms. For completeness, a structured summary of the main architectural features and loss components of each GAN family is available at the following link: \url{https://github.com/arco-group/MRI_TO_CT.git}.

\subsection{Training Protocol}

All ten architectures were trained using five-fold cross-validation at the patient level to prevent data leakage.

In each fold, 20\% of patients were used for testing, while the remaining 80\% were further split into training (64\%) and validation (16\%) subsets.
Each model was evaluated under seven training–testing configurations to analyze district-specific performance and cross-district generalization: AB, HN, and TH (training and testing on the respective single-district datasets); ALL AB, ALL HN, and ALL TH (training on the combined dataset ALL = AB + HN + TH and testing on each corresponding single-district dataset); and ALL (training and testing on the combined dataset ALL).
All models were trained for a maximum of 300 epochs with a 50 epoch warm-up phase. Early stopping with a patience of 50 epochs was applied based on validation loss.
Due to heterogeneous computational requirements, models were trained on different GPUs (NVIDIA T4, A100, V100, and A40).

\subsection{Evaluation Metrics and Comparative Analysis}

Synthetic CT images were quantitatively compared with the corresponding real CT images using complementary metrics capturing voxel-wise accuracy, structural fidelity, perceptual similarity, and distribution-level realism: (i) \textbf{Mean Squared Error (MSE)} – evaluates voxel-wise HU fidelity;
(ii) \textbf{Peak Signal-to-Noise Ratio (PSNR)} – measures global reconstruction quality;
(iii) \textbf{Structural Similarity Index Measure (SSIM)} – assesses structural and anatomical preservation;
(iv) \textbf{Visual Information Fidelity (VIF)} – quantifies preserved visual information in a multi-scale framework;
(v) \textbf{Learned Perceptual Image Patch Similarity (LPIPS)} – evaluates perceptual similarity in deep feature space;
(vi) \textbf{Fréchet Inception Distance (FID)} – measures distribution-level realism between real and synthetic CT images. Voxel-wise accuracy was assessed using MSE and PSNR, structural fidelity through SSIM and VIF, and perceptual similarity and distribution-level realism through LPIPS and FID. Interested readers can refer to \cite{di2023comparative} for their definition omitted here for space reasons.
All metrics were computed per slice and aggregated at the patient level before cross-fold averaging to avoid bias due to variable slice counts.
For each model, mean performance across folds and standard error of the mean were reported.

To offer a controlled comparison in terms of accuracy, realism, robustness, and cross-anatomical generalization, we systematically analyzed:

\begin{itemize}
\item  Comparison of the 10 networks by evaluating each model under seven train–test configurations (AB, HN, TH, ALL AB, ALL HN, ALL TH, ALL);
\item Learning paradigm (Paired vs Unpaired); 
\item Architectural family comparisons;
\item Cross training-test configurations (single-districts, ALL to single, ALL to ALL);
\item The average translation quality by aggregating networks into four categories based on the training–testing district (AB, HN, TH, ALL).
\end{itemize}

\section{Results and Discussion}

\subsection{Evaluation}

The quantitative evaluation analyses MRI--CT synthesis performance across different generative models, anatomical districts, and architectural families.

As a first analysis, models were compared within each anatomical district using the \textbf{Mean Rank}, defined as the average of the ranks obtained across all quantitative metrics (MSE, PSNR, VIF, SSIM, LPIPS, FID). In this context, the rank indicates the position assumed by a model for a given metric within a specific district, relative to the other models evaluated in the same setting. It is defined as: 

\begin{equation}
\text{MeanRank}_m = \frac{1}{6} \sum_{k=1}^{6} \text{rank}_{m,k},
\label{meanrank}
\end{equation}
where $\text{rank}_{m,k}$ represents the rank of model $m$ with respect to metric $k$.
The best value that the Mean Rank can assume is equal to 1 and occurs when the considered model ranks first for each of the evaluated metrics. When comparing two models, the more performant one is the model with the lower Mean Rank, as this value reflects a higher and more balanced overall performance across the different metrics.
Ties are resolved by computing the number of metrics for which the model achieves the best value.
It's worth noting that such a formulation emphasizes balanced performance across metrics and reduces sensitivity to any single metric. The results in Table \ref{tab:Summery_of_best_performing} show that UNIT achieves the best performance in the configurations AB and TH, standing out particularly in structural fidelity and perceptual quality, as highlighted by the VIF and LPIPS values reported in the Best metrics column, which includes the metrics with $\text{rank}_{m,k}$ = 1. Conversely, Pix2Pix dominates the multi-district scenarios (ALL *), where it attains the best Mean Rank, emerging as the top-performing model in up to five metrics in the “ALL AB” scenario and confirms its position as the most balanced model even in complex configurations such as “ALL”, where, although not first in any single metric, it shows the best overall balance across all metrics.

\begin{table}[t]
\raggedright
\caption{Summary of best-performing models across anatomical districts, supervision paradigms, and architectural families.}
\label{tab:Summery_of_best_performing}

\centering
\footnotesize
\setlength{\tabcolsep}{3pt}
\renewcommand{\arraystretch}{1.05}

\begin{tabular}{p{1.6cm} p{2.0cm} p{1.4cm} p{2.6cm}}
\hline
\textbf{Configuration} & \textbf{Best Model} & \textbf{Mean Rank} & \textbf{Best Metrics} \\
\hline
\multicolumn{4}{p{7.6cm}}{\textit{Best performing model per anatomical district}} \\
\hline
AB      & UNIT     & 2.00 & VIF, LPIPS \\
HN      & Pix2Pix  & 2.17 & PSNR, VIF, SSIM \\
TH      & UNIT     & 2.33 & VIF \\
ALL AB & Pix2Pix  & 1.67 & MSE, PSNR, VIF, SSIM, LPIPS \\
ALL HN & Pix2Pix  & 2.00 & VIF, SSIM, LPIPS \\
ALL TH & Pix2Pix  & 2.17 & VIF, SSIM, LPIPS \\
ALL     & Pix2Pix  & 2.67 & -- \\
\midrule
\multicolumn{4}{p{7.6cm}}{\textit{Paired vs Unpaired models}} \\
\hline
AB      & Paired Models & 1.00 & MSE, PSNR, VIF, SSIM, LPIPS, FID \\
HN      & Paired Models & 1.00 & MSE, PSNR, VIF, SSIM, LPIPS, FID \\
TH      & Paired Models & 1.00 & MSE, PSNR, VIF, SSIM, LPIPS, FID \\
ALL AB & Paired Models & 1.00 & MSE, PSNR, VIF, SSIM, LPIPS, FID \\
ALL HN & Paired Models & 1.00 & MSE, PSNR, VIF, SSIM, LPIPS, FID \\
ALL TH & Paired Models & 1.00 & MSE, PSNR, VIF, SSIM, LPIPS, FID \\
ALL     & Paired Models & 1.00 & MSE, PSNR, VIF, SSIM, LPIPS, FID \\
\midrule
\multicolumn{4}{p{7.6cm}}{\textit{Architectural families comparison}} \\
\hline
AB      & Paired Family & 1.17 & MSE, PSNR, VIF, SSIM, LPIPS \\
HN      & Paired Family & 1.17 & MSE, PSNR, VIF, SSIM, LPIPS \\
TH      & Paired Family & 1.17 & MSE, PSNR, VIF, SSIM, LPIPS \\
ALL AB & Paired Family & 1.17 & MSE, PSNR, VIF, SSIM, LPIPS \\
ALL HN & Paired Family & 1.17 & MSE, PSNR, VIF, SSIM, LPIPS \\
ALL TH & Paired Family & 1.17 & MSE, PSNR, VIF, SSIM, LPIPS \\
ALL     & Paired Family & 1.17 & MSE, PSNR, VIF, SSIM, LPIPS \\
\midrule
\multicolumn{4}{p{7.6cm}}{\textit{Global average across all districts}} \\
\hline
Global & Pix2Pix & 2.00 & PSNR, VIF, SSIM, LPIPS \\
\hline
\end{tabular}
\end{table}

The same Table \ref{tab:Summery_of_best_performing} permit us to investigate also the impact of the supervision paradigm and architectural family. The distinction between Paired and Unpaired models shows that Paired models consistently achieve superior performance across all considered scenarios, both single-district and multi-district. As observed, Paired models outperform Unpaired ones on every evaluated metric ($\text{rank}_{m,k}$ = 1), resulting in a Mean Rank = 1 in each district.
The same evidence emerges when aggregating the architectures into the five families reported in Section \ref{Gan Architectures}, i.e., Paired, Cycle-Based, Latent Space Sharing, Multi-domain, and Domain Adaptation. Also in this case, the Paired family prevails across all districts and training configurations, in line with the robustness observed for Pix2Pix. This conclusion is further supported by the global analysis, in which Pix2Pix emerges as the overall best-performing model with a Mean Rank equal to 2.00, achieving the best results in PSNR, VIF, SSIM, and LPIPS, as reported in the last section of Table \ref{tab:Summery_of_best_performing}. 
The third analysis examined three aggregation schemes of the training–testing configurations, obtained by averaging performance across all models. Specifically, the first section of Table \ref{tab:global_districts} reports results grouped into three macro-categories: (i) models trained and evaluated on individual district datasets (AB, HN, TH); (ii) models trained on the combined dataset and evaluated on each individual district (ALL AB, ALL HN, ALL TH); and (iii) models trained and evaluated on the combined dataset (ALL).
\begin{table}[t]
\raggedright
\caption{Average performance comparison across anatomical districts and training configurations.}
\label{tab:global_districts}

\centering
\footnotesize
\setlength{\tabcolsep}{4pt}
\renewcommand{\arraystretch}{1.05}

\begin{tabular}{p{3.8cm} p{1.0cm} p{3.2cm}}
\hline
\textbf{Configuration} & \textbf{Mean Rank} & \textbf{Best Metrics} \\
\hline
\multicolumn{3}{p{8.0cm}}{\textit{Across aggregated anatomical settings}} \\
\hline
ALL AB+ALL HN+ALL TH & 1.67 & VIF, SSIM, LPIPS \\
ALL                      & 2.00 & FID \\
AB+HN+TH               & 2.33 & MSE, PSNR \\
\midrule
\multicolumn{3}{p{8.0cm}}{\textit{Across anatomical districts}} \\
\hline
TH  & 1.83 & MSE, PSNR, VIF \\
ALL & 2.17 & FID \\
HN  & 3.00 & SSIM, LPIPS \\
AB  & 3.00 & -- \\
\hline
\end{tabular}
\end{table}

\begin{figure*}[t]
    \centering
    \includegraphics[width=0.9\linewidth]{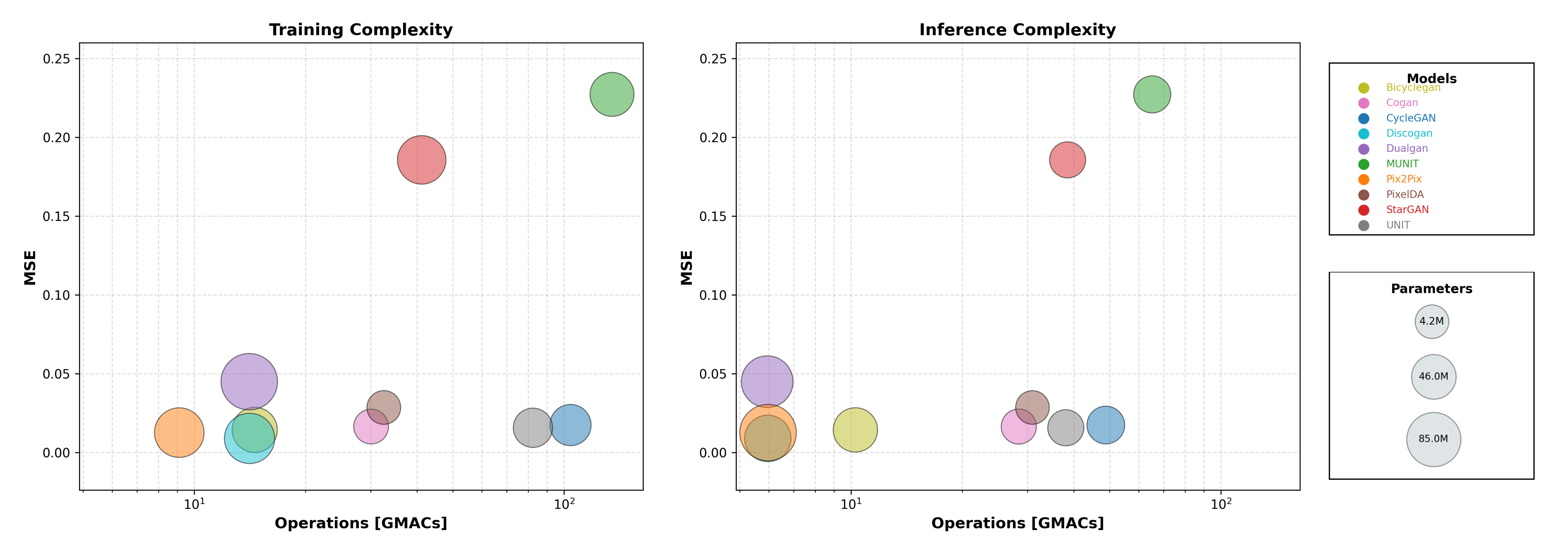}
    \caption{
    Computational complexity versus synthesis performance during training and inference.
    }
    \label{fig:bubbleplots}
\end{figure*}

\begin{figure*}[t]
    \centering
    \includegraphics[width=0.8\linewidth]{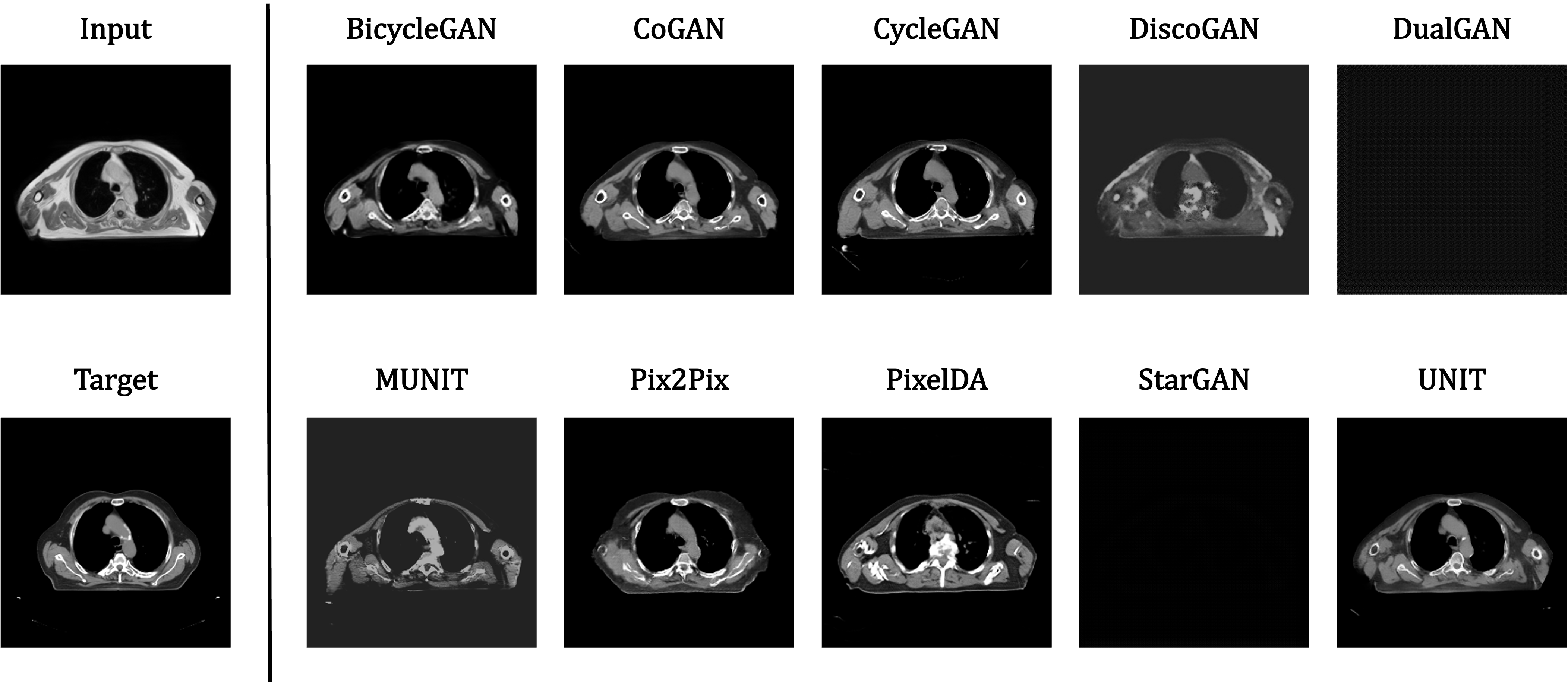}
    \caption{
    Qualitative comparison of synthetic CT images generated by ten GAN architectures.
    }
    \label{fig:output_model}
\end{figure*}

From the comparison among these configurations, it emerges that the ALL AB + ALL HN + ALL TH category achieves the best Mean Rank (1.67), dominating structural and perceptual metrics (VIF, SSIM, LPIPS). This result suggests that training on heterogeneous anatomical data may act as a regularization mechanism, improving robustness and perceptual stability during district-specific evaluation.
Conversely, the AB + HN + TH configuration achieves higher voxel-wise accuracy (MSE and PSNR), but a lower overall ranking. A structural trade-off therefore emerges: intra-district training maximizes local intensity fidelity, whereas multi-district training enhances generalization and perceptual consistency.

Let us now focus on the analysis per anatomical regions (second section of Table \ref{tab:global_districts}). The thoracic district (TH) achieves the best Mean Rank (1.83), dominating MSE, PSNR, and VIF. This behavior is consistent with the anatomical characteristics of the thorax, where high-contrast structures such as air cavities and bone interfaces reduce ambiguity in the MRI$\rightarrow$CT mapping. To ensure the reproducibility of our experiments we make our code public, together with the overall results, at the following link: \url{https://github.com/arco-group/MRI_TO_CT.git}

\subsection{Computational Complexity Evaluation}
Beyond synthesis performance, we evaluate the computational complexity in terms of number of parameters and operations (GMACs) during both training and inference. Figure~\ref{fig:bubbleplots} summarizes this comparison through bubble plots, where: the $x$-axis represents operations (GMACs), the bubble size is proportional to the number of parameters and the $y$-axis represents the value of the considered metric.
The complexity analysis considers two complementary aspects: training and inference. Training complexity encompasses the entire optimization process and is therefore influenced by the overall architectural design. Models composed of multiple interconnected subnetworks, such as dual generators and discriminators, encoder–decoder modules, or auxiliary classifiers, incur substantially higher GMACs and parameter counts, as several components must be jointly optimized during learning.
Inference complexity, by contrast, reflects only the forward pass required to generate the synthetic image. As a result, the computational burden is significantly reduced compared to training, particularly in architectures where inference relies on a single generator pathway. A direct comparison between training and inference reveals that computational cost does not scale uniformly across models. Multi-component architectures are considerably more demanding during training, while inference costs tend to converge once only one generative branch is active.
In line with the quantitative performance results, Paired models, and especially Pix2Pix, also demonstrate a favorable balance from a computational perspective. Their relatively streamlined architecture enables a more controlled trade-off between synthesis quality and computational efficiency, particularly during inference, where resource demands remain comparatively moderate.

\subsection{Qualitative Evaluation}

A visual comparison is shown in Figure~\ref{fig:output_model}, where representative thoracic slices are generated by all ten models using the same MRI input and CT reference. The thoracic district was selected due to its comparatively stable quantitative performance. Pix2Pix produces visually realistic and structurally coherent reconstructions, in agreement with its strong quantitative metrics. High PSNR reflects voxel-wise fidelity, while elevated VIF and SSIM confirm preservation of visual and anatomical structures. UNIT follows closely, producing stable anatomical representations with slightly smoother contours, reflecting a minor compromise in sharpness. CycleGAN and CoGAN generate plausible structures but exhibit mild smoothing and reduced definition, consistent with their relatively lower SSIM and PSNR values. PixelDA preserves global organization but shows less stable intensity consistency. In contrast, MUNIT and DiscoGAN display visible degradation in image quality, including contrast attenuation or localized artifacts, coherently reflecting their less favorable MSE, SSIM, and LPIPS performance. Notably, DualGAN and StarGAN consistently fail to produce anatomically reliable reconstructions across all evaluated configurations, as clearly illustrated in Figure~\ref{fig:output_model}, confirming their unstable quantitative behavior.

\section{Conclusion}

This study presented a systematic benchmark of ten GAN-based architectures for MRI-to-CT synthesis, evaluated across multiple anatomical districts and training paradigms under a unified 5-fold cross-validation protocol. Performance was assessed using complementary metrics capturing voxel-wise accuracy (MSE, PSNR), structural fidelity (SSIM, VIF), perceptual quality (LPIPS), and distribution-level realism (FID), alongside an analysis of computational complexity during training and inference.
The results demonstrate that Paired (supervised) models consistently outperform Unpaired approaches, highlighting the importance of voxel-wise supervision in reducing ambiguity in the MRI$\rightarrow$CT mapping and improving Hounsfield Unit consistency. Among the analyzed architectures, Pix2Pix emerged as the most robust and well-balanced model across districts and metrics, while maintaining a favorable trade-off between synthesis quality and computational cost. 
The results further indicate that multi-district training configurations lead to overall superior performance compared to district-specific ones, suggesting a regularization effect arising from exposure to greater anatomical variability.
Future work will extend this study in three directions. First, a clinically oriented evaluation will be introduced through a structured Turing test involving expert radiologists to assess perceptual realism and diagnostic usability. Second, alternative generative paradigms, including diffusion-based and transformer-based models, will be explored to investigate potential improvements in structural fidelity and generalization. Third, the benchmark will be expanded to additional anatomical districts and more heterogeneous acquisition protocols to further evaluate robustness in realistic clinical scenarios.

\section*{Acknowledgment}

Alessandro Pesci is a Ph.D. student enrolled in the National Ph.D. in Artificial Intelligence, XL cycle, course on Health and Life Sciences, organized by Università Campus Bio-Medico di Roma.
This work was partially funded by:
(i) Cancerforskningsfonden Norrland project (LP25-2383); 
(ii) Kempe Foundation project (JCSMK24-0094);
(iii)Università Campus Bio-Medico di Roma under the programme ‘‘University Strategic Projects’’ within the project ‘‘AI-powered Digital Twin for next-generation lung cancEr cAre (IDEA)’’; 
(iv) PNRR M6/C2 project PNRR-MCNT2-2023-12377755.

\bibliographystyle{IEEEtran}
\bibliography{references.bib}

\begin{comment}

\vspace{12pt}
\color{red}
IEEE conference templates contain guidance text for composing and formatting conference papers. Please ensure that all template text is removed from your conference paper prior to submission to the conference. Failure to remove the template text from your paper may result in your paper not being published.
\end{comment}

\end{document}